\documentclass[12pt,letterpaper]{article}
\usepackage[a4paper, total={7in, 10in}]{geometry}

\usepackage{graphicx}
\usepackage{helvet}
\usepackage{authblk}
\usepackage{hyperref}
\usepackage{amsmath} 
\usepackage{amssymb} 
\usepackage{orcidlink} 
\usepackage[super,comma,sort&compress]  
   {natbib}
\usepackage[right]{lineno} \linenumbers
\usepackage{subcaption}

\makeatletter
\renewcommand{\maketitle}{\bgroup\setlength{\parindent}{0pt}
\begin{flushleft}
  \textbf{\@title}
  
  \@author
\end{flushleft}\egroup}
\makeatother

\title{Self-Aligning EPM Connector: A Versatile Solution for Adaptive and Multi-Modal Interfaces}
\date{}

\author[1,*] {Bingchao Wang}
\author[1, 2, 3, 4, *] {Adam A. Stokes}

\affil[1]{Soft Systems Group, The School of Engineering, The University of Edinburgh, EH9 3FF Edinburgh, UK}
\affil[2]{Senior author}
\affil[3]{X (formerly Twitter): @adamastokes}
\affil[4]{Lead contact}

\affil[*]{Correspondence: B.W. B.Wang-57@sms.ed.ac.uk}
\affil[*]{Correspondence: A.A.S. adam.stokes@ed.ac.uk}

\begin{document}

\maketitle

\section*{SUMMARY}

This paper introduces an advanced multifunctional connector utilizing electro-permanent magnet (EPM) technology designed to address limitations in a variety of commercial and industrial applications. The proposed connector integrates self-alignment, mechanical coupling, fluid transfer, and robust data communication within a compact structure fabricated via SLA-based 3D printing. Experimental analyses demonstrate the connector's high reliability in mechanical alignment, efficient fluid transfer under both single-loop and dual-channel modes, and effective data transmission through integrated electronic control systems. Furthermore, from a conceptual perspective, the connector architecture supports notable mechanical flexibility, accommodating significant axial extension, angular misalignment, lateral offset, and extended connection distances, applicable to different structural implementations. The optimized EPM design, incorporating a coil wound exclusively around the Alnico magnet, significantly enhances magnetic performance and energy efficiency. These features collectively ensure robust and versatile functionality, positioning this connector as highly suitable for modular robotics, electric vehicle charging ports, domestic robotic platforms, and aerospace applications requiring precise, adaptable interconnections.

\section*{KEYWORDS}


EPM, Connector, Self-alignment, Fluid transfer

\section*{INTRODUCTION}

In modular robotics and numerous everyday systems, performance effectiveness increasingly relies on innovative connector technologies. Connectors designed for self-alignment facilitate rapid reconfiguration, robust integration, and adaptability, enabling systems to swiftly respond to dynamic or unstructured environments. Modular robots particularly benefit from these connectors, as they allow flexible assembly, simplified maintenance, and reliable module integration, essential for adapting to complex tasks and uncertain conditions \cite{yim2007modular,zhang2020modular,saab2019review,seo2019modular}. Beyond robotics, versatile self-aligning connectors play critical roles in everyday applications, including electric vehicle (EV) charging interfaces, automated vacuum docking stations, and integrated smart home systems. Such connectors significantly enhance docking efficiency, minimize manual intervention, and improve reliability during daily operations \cite{song2011automatic}. Moreover, in challenging environments such as space exploration, reliable self-aligning connectors facilitate spacecraft docking, satellite servicing, and the modular assembly of space stations, demanding precise autonomous operations under constrained and unpredictable conditions \cite{flores2014review,papadopoulos2021robotic}. As systems become more interconnected and multifunctional, connectors capable of supporting simultaneous electrical, data, and fluidic transfers are increasingly essential \cite{arif2021review}.\\
Controllable connection mechanisms are typically categorized as mechanical, magnetomechanical, or electromagnetic \cite{eckenstein2014area}. Traditional module interconnections generally utilize mechanical latching mechanisms due to their reliability in ensuring secure coupling. Representative examples include the HiGen connector, featuring four radial hooks that rotate and mechanically interlock with complementary hooks, achieving strong and repeatable connections \cite{parrott2014hiclaws}. Additionally, the SINGO Connector employs an actively driven spiral gear mechanism, providing substantial alignment tolerance and error correction capabilities \cite{shen2009singo}. The X-Claw connector, another notable mechanical design, uses an active gripping mechanism combined with self-alignment features, enabling high acceptance angles and multiple attachment orientations between modules \cite{cong2011xclaw}. Despite their reliability and stability, mechanical connectors' dependence on motors and active components introduces spatial constraints, limiting their applicability in compact or resource-constrained environments.\\
To address some of these limitations, many connector designs incorporate permanent magnets for effective alignment and coupling \cite{zhao2022soft,vergara2017soft}. For instance, the EMERGE Modular Robot integrates magnetic connectors directly into module mating surfaces, enabling efficient alignment and secure attachment \cite{moreno2021emerge}. Similarly, robotic units equipped with ring-shaped permanent magnets facilitate precise self-alignment during connection \cite{lee2017development}. Although effective for alignment, these permanent magnet-based connectors inherently lack active self-detachment capabilities, restricting their versatility in dynamic and reconfigurable scenarios.\\
Electro-permanent magnets (EPMs) have increasingly become the connector choice due to their rapid switching capability, energy efficiency, and ability to maintain strong connections without continuous power consumption \cite{tapus2007socially,romanishin2013mblocks}. Prominent examples include the EP-Face connector, employing a planar array of EPMs for rapid switching and robust holding strength, significantly enhancing module connectivity performance \cite{tosun2016design}. Similarly, the SMORES-EP system uses EPM-based connectors on multiple faces, facilitating versatile and reliable robotic reconfigurations and highlighting advantages in modular flexibility and scalability \cite{davey2012emulating}. Another notable approach is the M-Blocks robotic modules, which leverage EPMs to enable self-assembly and autonomous mobility by dynamically adjusting magnetic polarity for controlled attachment and detachment \cite{gilpin2010robot,daudelin2018integrated}. Despite these advancements, current EPM-based connectors still exhibit limitations, such as reduced performance under shear and torsional loads \cite{tosun2016design}, and insufficient multifunctionality—particularly regarding fluidic or hybrid connections \cite{davey2012emulating}. Consequently, significant potential exists for developing enhanced EPM connectors that combine multifunctionalcoupling and mechanical robustness.
Recent advances in modular robotic connectors emphasize integrating fluidic, electrical, and communication functionalities within a unified interface \cite{bray2023recent}. Several standardized interfaces incorporating fluid or air transfers have emerged to enable these functionalities effectively. The PAC (Power, Air, Communication) connector facilitates seamless resource sharing, combining pneumatic transfer capabilities with robust mechanical and magnetic locking mechanisms for secure and reversible module connections \cite{knospler2024shared}. Another innovation is the Standard Interface for Robotic Manipulation (SIROM), developed by SENER Aeroespacial, integrating mechanical coupling, electrical interfaces, refueling, thermal transfer, and data communication into a compact, unified solution \cite{diaz2023sirom,sirom_sener,guerra2022development}. Similarly, GMV’s ASSIST system employs a dedicated berthing fixture with integrated fluid connectors, facilitating zero-force docking and secure fluid transfer between spacecraft \cite{medina2017towards,guerra2022development}. However, these connector designs generally lack intrinsic self-alignment capabilities, relying instead on external systems or precise manual alignment, significantly limiting their effectiveness in dynamic or autonomous scenarios, where rapid and accurate self-alignment is crucial \cite{eckenstein2014area}. Orbit Fab's Rapidly Attachable Fluid Transfer Interface (RAFTI) employs alignment markers known as fiducials to facilitate docking by providing visual cues during proximity operations, thus reducing docking complexity \cite{orbitfab_rafti}. Despite improving docking precision, visual-based alignment systems like RAFTI's fiducials possess inherent limitations, particularly in low-visibility conditions such as poor lighting or obstructed views, potentially hindering accurate visual detection and alignment \cite{bian2023fima,alijani2017autonomous}.\\
To address these limitations, we introduce a novel self-aligning connector incorporating an electro-permanent magnet (EPM) system (Fig.~1) to achieve multifunctionality in mechanical connection, fluid transfer, and data communication. Our proposed connector architecture is designed specifically to overcome existing limitations, targeting modular soft robotic systems and extending to broader commercial applications, including EV charging ports and domestic robotic systems such as household mopping robots. Key features of our design include: (i) a compact and integrated mechanical structure fabricated via SLA-based 3D printing technology, equipped with EPMs, bearings, springs, and other components for robust and reliable mechanical connections; (ii) embedded electronic control provided by a custom-designed PCB with an integrated STM32 microcontroller, facilitating robust data communication via UART; (iii) an internal isolation design ensuring complete sealing of internal mechanical and electronic components from transported fluids; (iv) mechanical decoupling capability enabling relative rotation, precise self-alignment, and accommodation of minor positional misalignments between connected modules; (v) low-energy EPM activation requiring only 0.3 J per switching event—achieved via a 30 V, 10 A pulse lasting 1 ms (Fig.~S2, Fig.~S3)—supporting energy-efficient operation; and (vi) versatile fluid transfer functionality capable of both unidirectional and bidirectional fluid exchange, applicable to modular robotics, EV charging systems, household robotics, and space applications requiring reliable fluid management and robust mechanical interconnections.
\begin{figure}[h] 
    \centering
    \includegraphics[width=0.6\columnwidth]{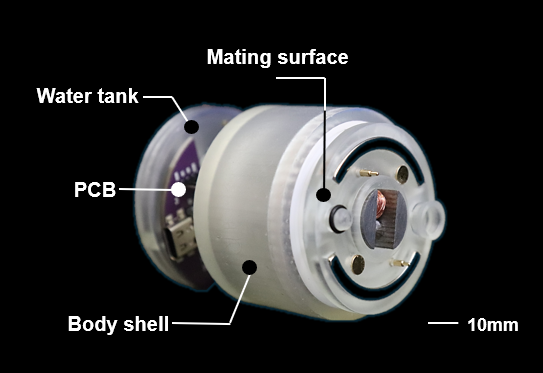} 
    \caption{Integrated Structure of Self-Aligning EPM Connector.}
    \label{fig:your_label}
\end{figure}

\begin{figure}[h] 
    \centering
    \includegraphics[width=0.8\columnwidth]{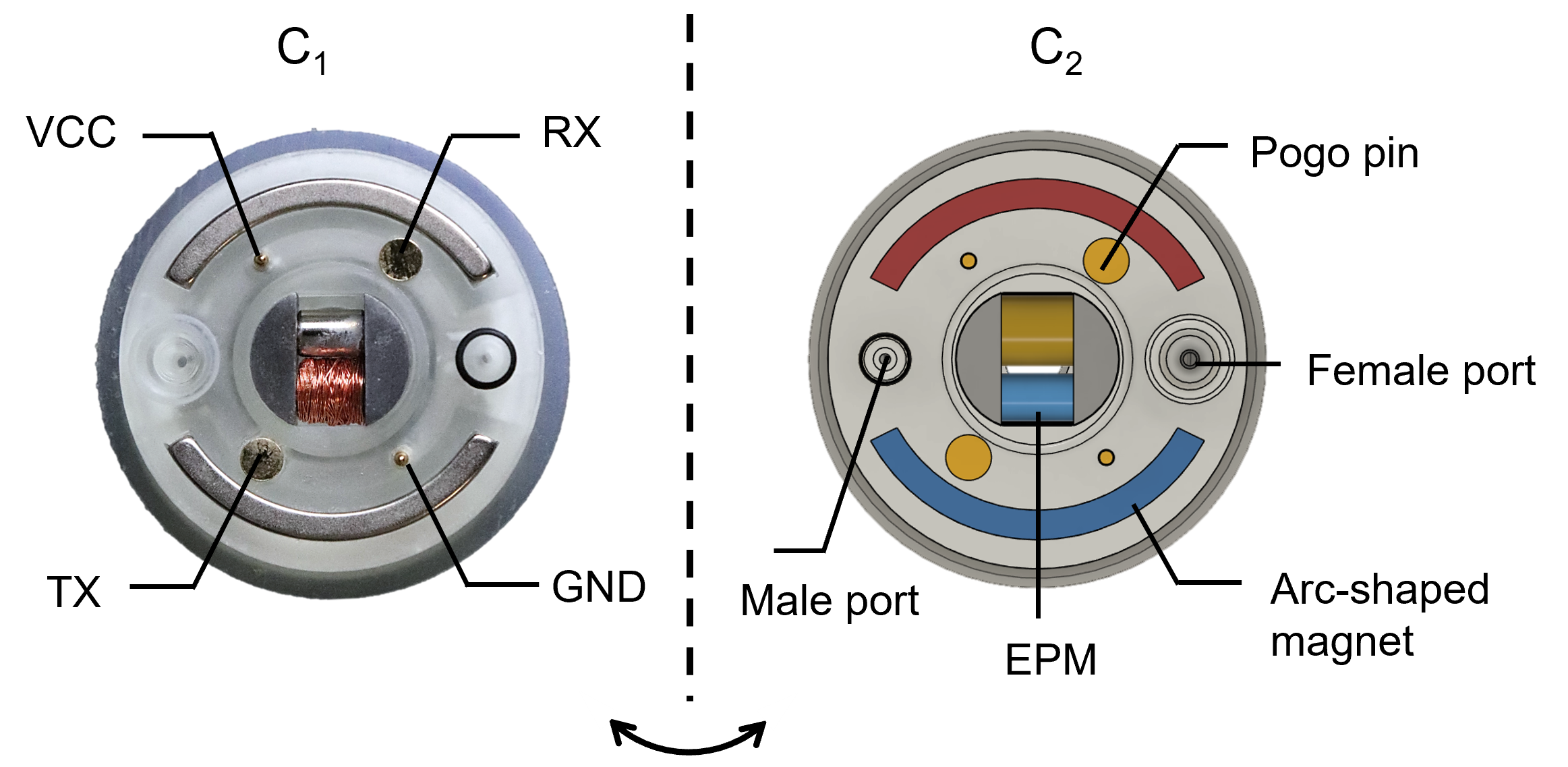} 
    \caption{Mating surface in EPM connector.}
    \label{fig:your_label}
\end{figure}

\section*{RESULTS}
\subsection*{EPM connector design}
The conception of the proposed Electro-Permanent Magnet (EPM) connector involves integrating two different pole arc-shaped permanent magnets and one electro-permanent magnet embedded within the mating surface (Fig.~2). As the two mating surfaces approach, the connector autonomously aligns through rotational movements enabled by integrated bearings, driven by the combined effects of magnetic attraction and repulsion, thus achieving precise self-alignment (Fig.~S4, Video S1). Within a certain proximity, the EPMs attract each other to firmly secure the mating surfaces. Consequently, the male port on the C	extsubscript{1} interface aligns and connects with the corresponding female port on the C	extsubscript{2} interface, and similarly, the male port on the C	extsubscript{2} interface connects to the female port on the C	extsubscript{1} interface, successfully establishing a reliable fluidic channel upon connector engagement. Simultaneously, the pogo pins on the mating surfaces establish one-to-one connections. Demagnetizing the EPMs results in the disconnection of the mating surfaces, separating the fluid channels and pogo pin connections (Fig.~S3).\\
\begin{figure}[!t] 
    \centering
    \includegraphics[width=0.9\columnwidth]{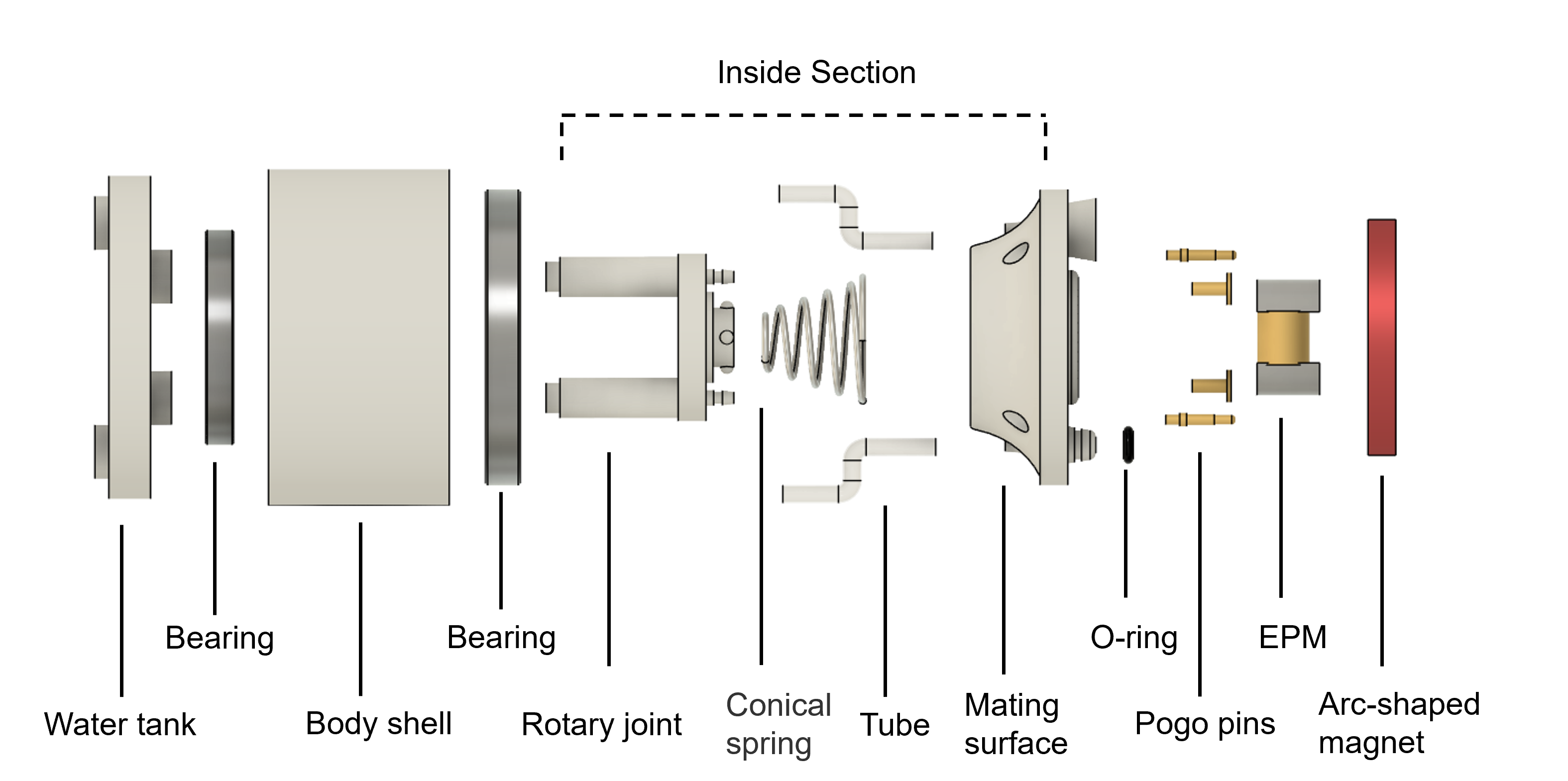} 
    \caption{Structure of EPM connector.}
    \label{fig:your_label}
\end{figure}\\
The EPM connector consists of four primary components: (i) an EPM, (ii) a conical spring (Metrol Ltd, UK), (iii) two bearings (Simply Bearings Ltd, UK), and (iv) a fluid channel system. Upon entering the water tank, fluid passes through a rotary joint and flows through a silicone tube into the internal section (Fig.~3), subsequently exiting via the mating surface. This constitutes the complete fluid channel system within the connector. The integration of the conical spring and bearings allows flexible coupling, accommodating angular misalignments up to 20 degrees (Fig.~4). Furthermore, the internal structure of the connector permits 360-degree rotation relative to the body shell, enabling synchronized rotation of both connectors upon engagement, significantly enhancing adaptability and robustness in dynamic operational environments.
\begin{figure}[!t] 
    \centering
    \includegraphics[width=0.5\columnwidth]{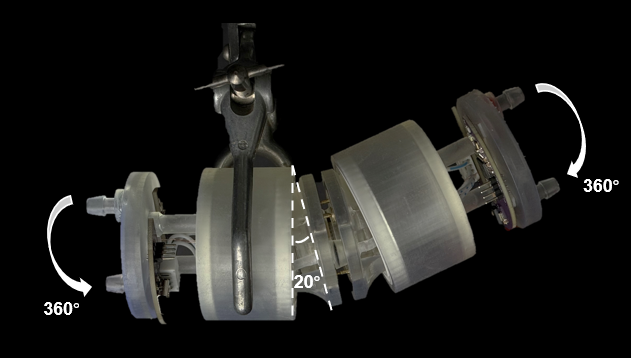} 
    \caption{EPM connector flexibility.}
    \label{fig:your_label}
\end{figure}
The water tank of each EPM connector includes two independent fluid inlets, internally forming isolated fluid transmission paths that allow for the simultaneous transfer of two different fluids without cross-contamination. When two EPM connectors are coupled, the system supports three distinct modes of fluid transfer (Fig.~5): (i) parallel unidirectional dual-channel fluidic transfer, where fluid flows simultaneously through two separate parallel channels from one connector into the other (Video S4); (ii) dual-channel fluidic transfer, where fluids flow simultaneously in opposite directions through two separate and isolated channels between connectors (Video S5); and (iii) single-loop fluidic transfer, in which fluid enters through one connector, passes into and through the other connector, and returns via a separate pathway, creating a closed-loop circuit (Video S6). Throughout all modes, the fluid pathways remain fully isolated to ensure fluid purity and prevent contamination from internal components. In the inside section, a conical spring connects the mating surface to the rotary joint, while bearings are positioned at both ends of the body shell—one mounted around the rotary joint and the other forming a line contact with the mating surface (Fig.~S7). The combined action of the conical spring and the bearings enables synchronized rotational motion among the mating surface, the rotary joint, and the water tank, while also allowing a degree of relative flexibility between the connected units (Video, S2). The fabrication of the water tank, body shell, rotary joint, and mating surface in the EPM connector is carried out using stereolithography (SLA) 3D printing technology.\\
\begin{figure}[!t] 
    \centering
    \includegraphics[width=0.6\columnwidth]{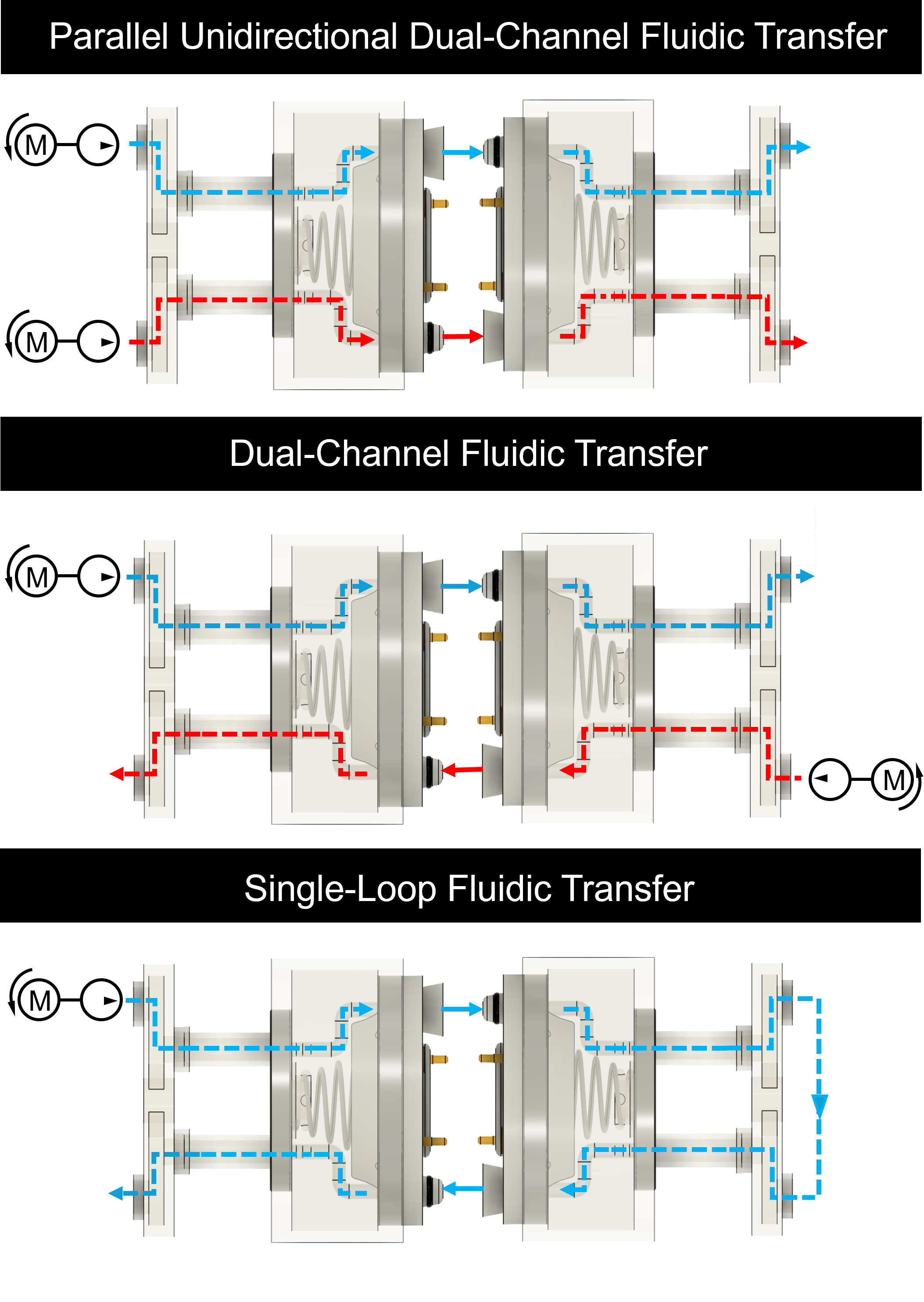} 
    \caption{Three models of fluid transfer in EPM connector.}
    \label{fig:your_label}
\end{figure}\\
The connectors are interconnected via pogo pins, providing both electrical power and UART-based data transmission, managed by onboard STM32 microcontrollers. The symmetrical arrangement of the pogo pins enables flexible connectivity in multiple orientations, enhancing adaptability during connector assembly. The mating surfaces contain pogo pins designated as VCC, GND, TX, and RX, which establish corresponding electrical connections when the connectors mate. The spring-loaded design ensures consistent electrical contact, facilitating stable and reliable data communication.Onboard LEDs, controlled by the STM32 microcontrollers and mounted on the rotating PCB located above the water tank, provide intuitive visual feedback regarding connection and communication status. A red LED briefly illuminates during active data transmission between connectors, clearly indicating ongoing communication. A green LED remains continuously illuminated upon successful mechanical and electrical connection, visually confirming stable connectivity. The PCB, along with the integrated STM32 controllers and LEDs, rotates synchronously with the water tank and pogo-pin connector assembly, maintaining reliable communication throughout dynamic operation (Fig.~S8, Video S3).

\begin{figure}[!t] 
    \centering
    \includegraphics[width=0.7\columnwidth]{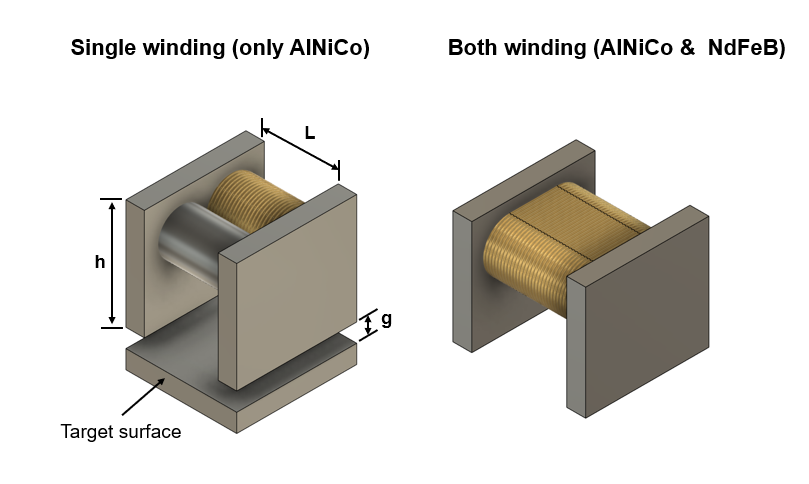} 
    \caption{Two coil winding methods of EPM.}
    \label{fig:your_label}
\end{figure}

\subsection*{EPM design and optimization}
An electro-permanent magnet (EPM) consists of two rod magnets with different coercivities: a hard magnet made of NdFeB and a soft magnet made of AlNiCo, along with end caps. A coil is positioned to generate a magnetic field that selectively changes the polarization of the soft magnet when pulsed with current. The magnetic force at the end caps is enabled when both magnets are polarized in the same direction and disabled when their polarizations oppose each other \cite{knaian2010electropermanent}.\\
Electro-permanent magnets (EPMs) have been extensively utilized across various fields, employing diverse fabrication methods. Generally, two common approaches exist for coil placement in EPM fabrication (Fig.~6). In the first method, the coil is wound around both AlNiCo and NdFeB magnets \cite{padovani2016electropermanent,nakayama2024deformable,kato2023switchable}. In the second method, the coil is wound exclusively around the AlNiCo magnet \cite{mcdonald2022modulation}. In the EPM connector, the force generated by the electro-permanent magnet serves as the coupling force between connectors. To estimate this force, we adopt the magnetic circuit model proposed by Knaian \cite{knaian2010electropermanent}, making several modifications to suit our specific application.\\
The general expression for the magnetomotive force (MMF) balance in the EPM magnetic circuit can be written as:
\begin{equation}
NI = 2H_g g + \sum_{i} \sigma_i H_i L_i\tag{1}
\end{equation}
Where $H_i$ and $L_i$ are the magnetic field intensity and effective length of each segment, respectively, and $\sigma_i \in \{+1, -1\}$ is a sign factor that depends on the relative orientation of the magnetization of the segment with respect to the reference direction of the magnetic circuit. A segment with $\sigma_i = +1$ acts as an MMF-consuming element (demagnetizing), while a segment with $\sigma_i = -1$ acts as an MMF-contributing element (magnetizing). This formulation highlights that the role of each magnet in the circuit is not fixed but depends on its polarization state relative to the circuit.\\
For the purpose of analysis, we first consider a representative case in which both the AlNiCo and NdFeB magnets act as MMF-consuming elements, i.e., $\sigma_{Al} = \sigma_{Nd} = +1$. Under this assumption, the effective MMFs available at the air gaps become:  
\begin{align}
\mathcal{F}_A &= NI - H_{\mathrm{Al}} L_{\mathrm{Al}} \tag{2} \\
\mathcal{F}_B &= NI - H_{\mathrm{Al}} L_{\mathrm{Al}} - H_{\mathrm{Nd}} L_{\mathrm{Nd}} \tag{3}
\end{align}
Where $\mathcal{F}_A$ corresponds to the effective MMF in the single-winding configuration (coil wound exclusively around the AlNiCo magnet), and $\mathcal{F}_B$ corresponds to the effective MMF in the both-winding configuration (coil wound around both the AlNiCo and NdFeB magnets). \\ 
Although the total MMF generated by the coil, $NI$, is identical in both cases, the utilization efficiency differs significantly. In configuration A, nearly all of the supplied MMF is directed toward the AlNiCo magnet, which has relatively low coercivity and can be effectively re-magnetized. In configuration B, however, part of the MMF is consumed within the NdFeB segment. Due to its high coercivity, this MMF does not contribute to switching and is effectively wasted, thereby reducing the usable MMF for re-magnetizing the AlNiCo magnet. This difference in MMF utilization underpins the superior efficiency of the single-winding strategy. \\ 
For convenience, the effective MMF delivered to the air gaps is denoted as $\mathcal{F}_{\mathrm{eff}}$, which takes the value of either $\mathcal{F}_A$ or $\mathcal{F}_B$ depending on the winding configuration. Once $\mathcal{F}_{\mathrm{eff}}$ is determined, the magnetic field intensity $H_g$ and the magnetic flux density $B_g$ in the air gaps can be calculated as:  
\begin{equation}
H_g = \frac{\mathcal{F}_{\mathrm{eff}}}{2g}, \quad
B_g = \mu_0 H_g
\tag{4}
\end{equation}
Where $H_g$ is the magnetic field intensity in the air gap (A/m), $\mathcal{F}_{\mathrm{eff}}$ is the effective MMF delivered to both air gaps (ampere-turns), $g$ is the air-gap thickness (m), $B_g$ is the resulting magnetic flux density (T), and $\mu_0 = 4\pi \times 10^{-7}\, \mathrm{H/m}$ is the permeability of free space. \\
This relationship clearly indicates that the magnetic flux density in the air gap is directly proportional to the effective MMF provided by the coil and inversely proportional to the total air-gap length. Thus, the coil winding configuration significantly influences the resultant flux density. Practically, increasing the coil turns while minimizing internal MMF losses within the magnet segments effectively enhances the magnetic field strength achievable in the air gap.\\
\begin{figure}[!t] 
    \centering
    \includegraphics[width=0.9\columnwidth]{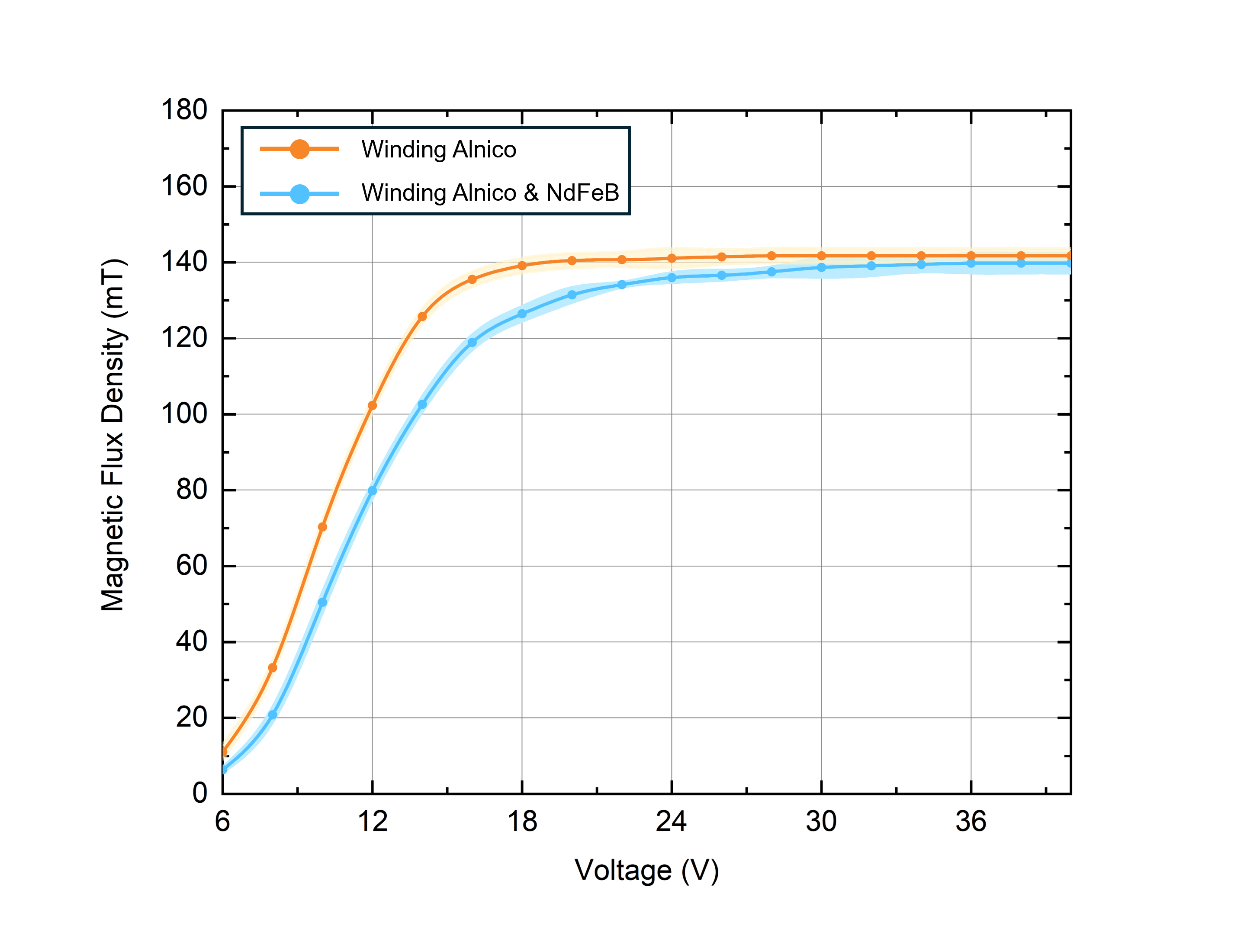} 
    \caption{Comparison of Magnetic Flux Density for Different Coil Winding Configurations.}
    \label{fig:your_label}
\end{figure}
\begin{figure}[!t] 
    \centering
    \includegraphics[width=0.5\columnwidth]{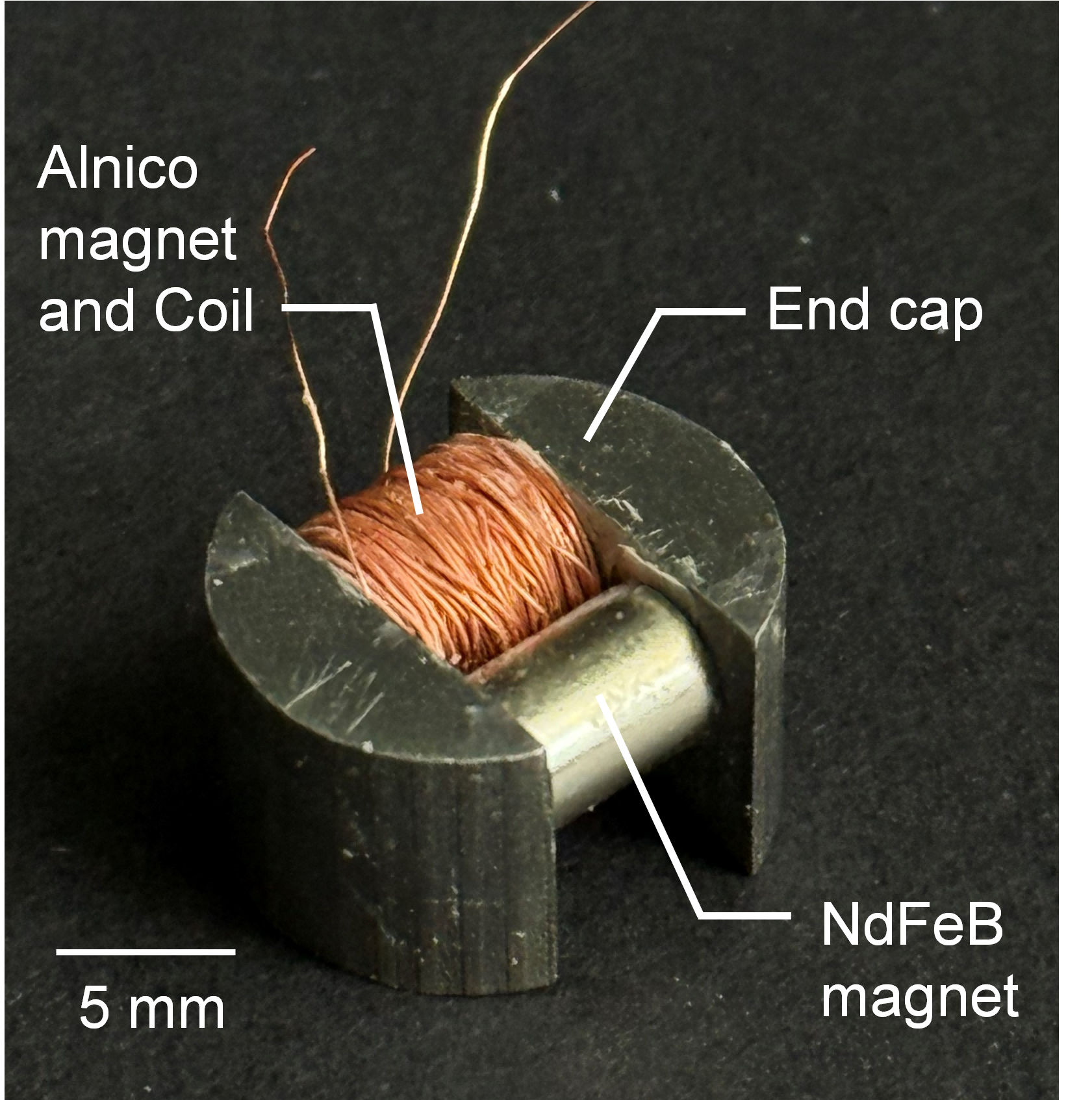} 
    \caption{EPM with ircular-segment-shape end caps. }
    \label{fig:your_label}
\end{figure}\\
To validate the theoretical predictions regarding the influence of coil winding strategies on magnetic flux density, systematic experiments were conducted using electro-permanent magnet (EPM) prototypes. Each prototype consisted of an AlNiCo 5 rod magnet (7 mm in length, 5 mm in diameter), a Grade N40 NdFeB disk magnet of identical dimensions, and precision-machined Q235 steel end caps (13 mm × 11 mm × 3 mm). Two coil configurations were fabricated: configuration A, in which a coil of 130 turns of 0.20 mm diameter copper wire was wound exclusively around the AlNiCo magnet, and configuration B, in which a coil of 130 turns of 0.22 mm diameter copper wire encompassed both the AlNiCo and NdFeB magnets. The coil resistances were adjusted to approximately $2.0 \, \Omega$ to ensure equivalent magnetomotive force ($NI$) under identical voltage conditions, thereby allowing direct and fair comparison between the two winding strategies.\\  
The experimental results (Fig.~7) demonstrate that the single-winding configuration consistently achieved higher magnetic flux density and reached saturation more rapidly across all tested voltage levels. By contrast, the both-winding configuration yielded lower flux density, as part of the coil MMF was expended across the NdFeB segment, whose high coercivity prevented effective re-magnetization. These findings confirm the theoretical analysis: although the total MMF is identical in both cases, the single-winding design utilizes it more efficiently by directing nearly all of the MMF toward switching the AlNiCo magnet. Consequently,the single-winding strategy was selected in this project as the more energy-efficient coil design for prototype development and further validation.\\
Based on these analytical and experimental findings, the EPM employed in the fluidic connector consisted of the aforementioned AlNiCo 5 rod magnet (7 mm × 5 mm, Guys Magnets Ltd, UK) and Grade N35 NdFeB disk magnet (Radial Magnets Ltd, US). A 120-turn coil fabricated from enamelled copper wire (35 AWG) with a diameter of 0.15 mm and polyurethane insulation was wound around the AlNiCo magnet (Fig.~8). The end caps, fabricated from Q235 steel, were designed in a circular-segment shape with an approximate area of 48.85 mm$^2$. This circular-segment shape was chosen to optimize the effective contact area between connectors, improving coupling efficiency relative to a rectangular geometry. The coil's electrical resistance was confirmed as 3.0 $\Omega$ using a multimeter. Detailed fabrication procedures are provided in the supplemental information (Fig.~S1).\\
To evaluate the practical normal force capability of the connector, experiments were conducted to investigate the impact of varying gap distances on the holding strength of EPMs. Two EPMs were arranged in a tensile testing configuration, with one magnet fixed and the other connected to a force gauge. Tensile force was incrementally applied until magnet separation occurred, recording the peak force values. Gap distances ranging from 0 mm to 1 mm were simulated using paper sheets in increments of 0.1 mm. Paper sheets were selected due to their magnetic reluctance properties closely resembling air, ensuring realistic experimental conditions. Before each measurement, EPMs were demagnetized and reactivated to maintain test consistency. Experimental results (Fig.~9) revealed a sharp decline in holding force, from approximately 14.6 N at zero gap to about 7.7 N at a minimal gap of 0.1 mm, nearly halving the force. Subsequently, the holding force gradually diminished, reaching approximately 2.34 N at the maximum tested gap of 1 mm. These results highlight the extreme sensitivity of EPM holding force to even minor gap increases. Further experimental details regarding EPM holding force are available in the supplemental information (Fig.~S5, Fig.~S6).

\begin{figure}[h] 
    \centering
    \includegraphics[width=1\columnwidth]{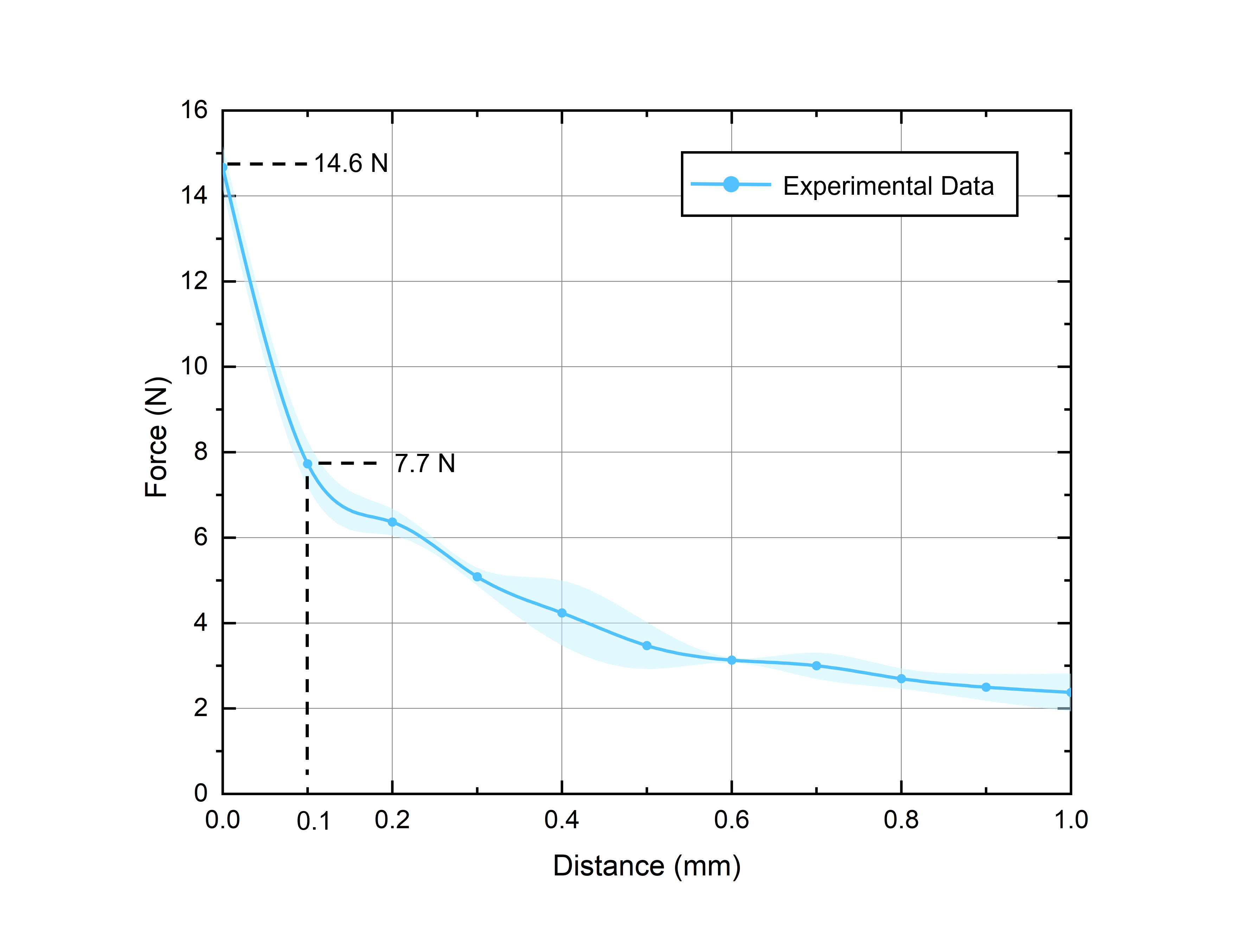} 
    \caption{Force vs. air gap for the EPM.}
    \label{fig:your_label}
\end{figure}

\subsection*{Self-Alignment Performance}
The connector employs magnetic attraction–repulsion, assisted by bearings, to achieve self-alignment. Its performance was evaluated using an improved method based on Zoltan’s approach \cite{nagy2007magnetic}. One connector (C\textsubscript{1}) was fixed on a platform with adjustable tilt angle ($\alpha$), while the other (C\textsubscript{2}) was lowered vertically at 3 mm/s from a height of 30 mm (Fig.~10). The test space was defined as a 7×7 grid with 5 mm increments in both x and y directions (Fig.~11).Two outcomes were observed: successful connections, where magnetic attraction overcame gravitational and elastic forces, and failed connections, where alignment could not be achieved. At $\alpha = 0^\circ$, connections were successful within 30 mm along the x-axis and 25 mm along the y-axis, giving a 59.1\% success rate. When tilted to $\alpha = 10^\circ$, the effective area reduced to 25 mm in both directions, lowering the success rate to 55.1\%. At $\alpha = 20^\circ$, connections were limited to 25 mm in x and 15 mm in y, with a further reduced success rate of 44.9\%.\\
These results indicate that although performance decreases with tilt, the connector still maintains robust self-alignment capability under moderate angular misalignment.

\begin{figure}[!t] 
    \centering
    \includegraphics[width=0.5\columnwidth]{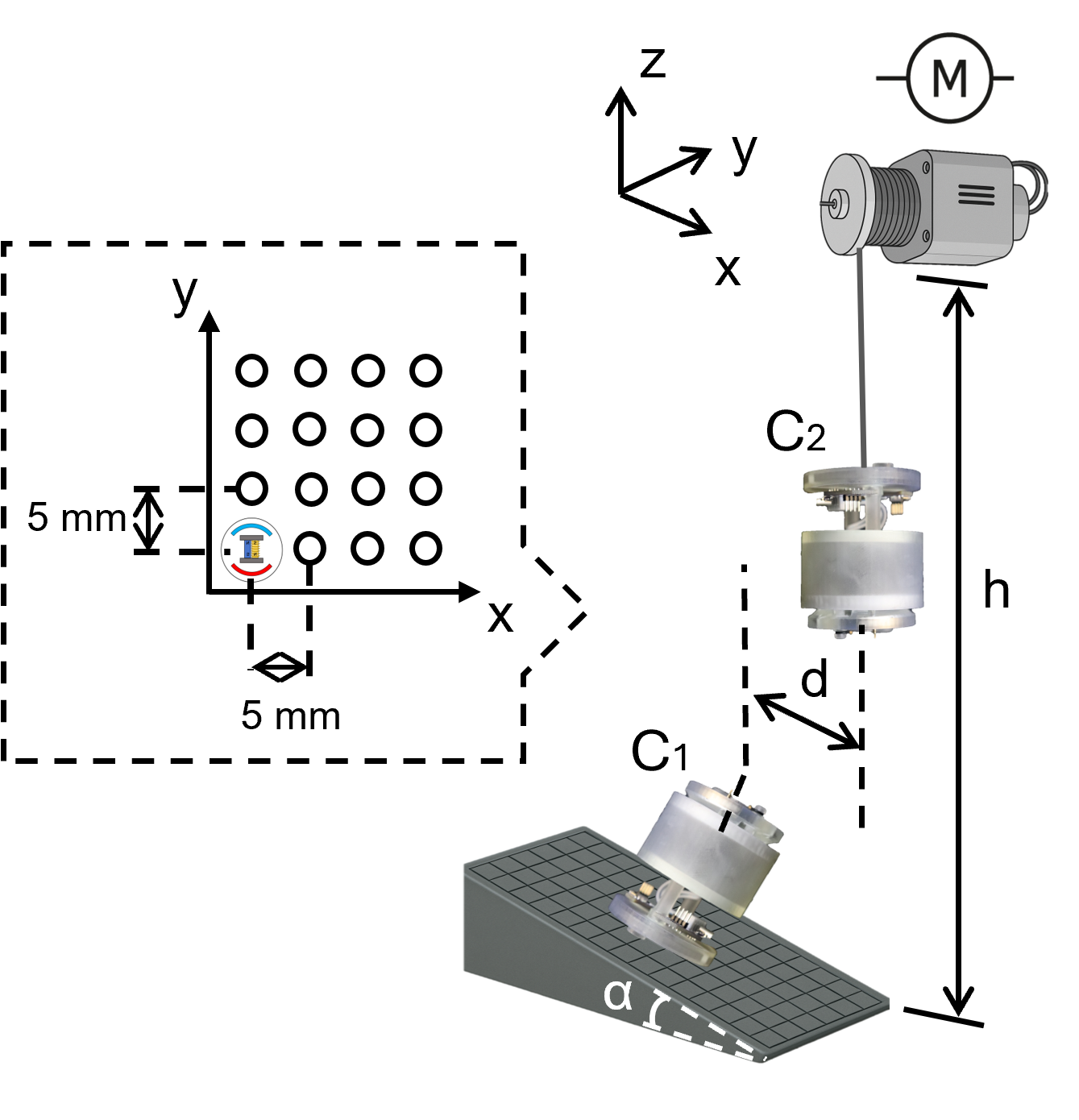} 
    \caption{Experimental sketch of self-alignment performance.}
    \label{fig:your_label}
\end{figure}
\begin{figure}[!t] 
    \centering
    \includegraphics[width=1\columnwidth]{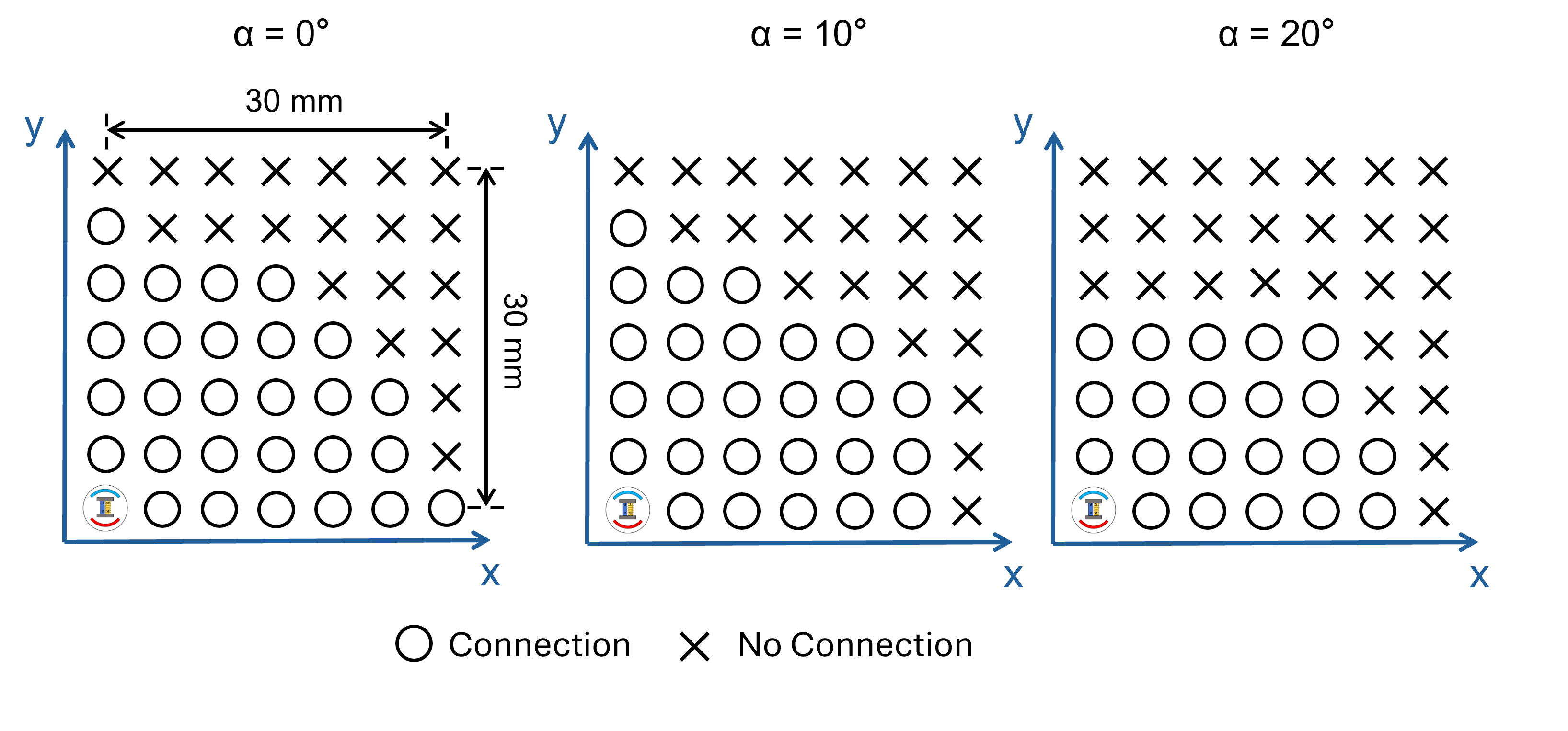} 
    \caption{Experimental results in parameter space.}
    \label{fig:your_label}
\end{figure}
\subsection*{Fluid transfer Performance}
Upon connection, the male fluidic ports on the mating surface of one connector engage precisely with the corresponding female ports of the other connector through magnetic self-alignment. Concurrently, an O-ring positioned on the male fluidic ports is compressed under the magnetic force generated by the electropermanent magnet (EPM), effectively sealing the interface between the male and female ports. This tight sealing mechanism ensures reliable fluid transfer by preventing leakage during operation.
To evaluate fluid transmission efficiency, flow rates at both the inlet and outlet of the connector were systematically measured. The experimental setup incorporated two Sensirion SLF3S-4000B (Sensirion, Switzerland) flow sensors positioned at the inlet and outlet, respectively. Driven by a water pump, fluid passed through the inlet flow sensor and entered the connector's water tank. It then flowed internally through the connector's channels, exited via the outlet flow sensor, and finally returned to the container (Fig.~12). Two distinct fluid transfer modes were examined: single-loop fluidic transfer and dual-channel fluidic transfer. For each mode, experiments were conducted at different input flow rate, enabling the collection and analysis of three sets of flow rate data per mode.Each test lasted two minutes, with 30 seconds of data randomly sampled for analysis. Before each test, the system's pipes were emptied to ensure the accuracy of the experimental results.\\
\begin{figure}[!t] 
    \centering
    \includegraphics[width=0.8\columnwidth]{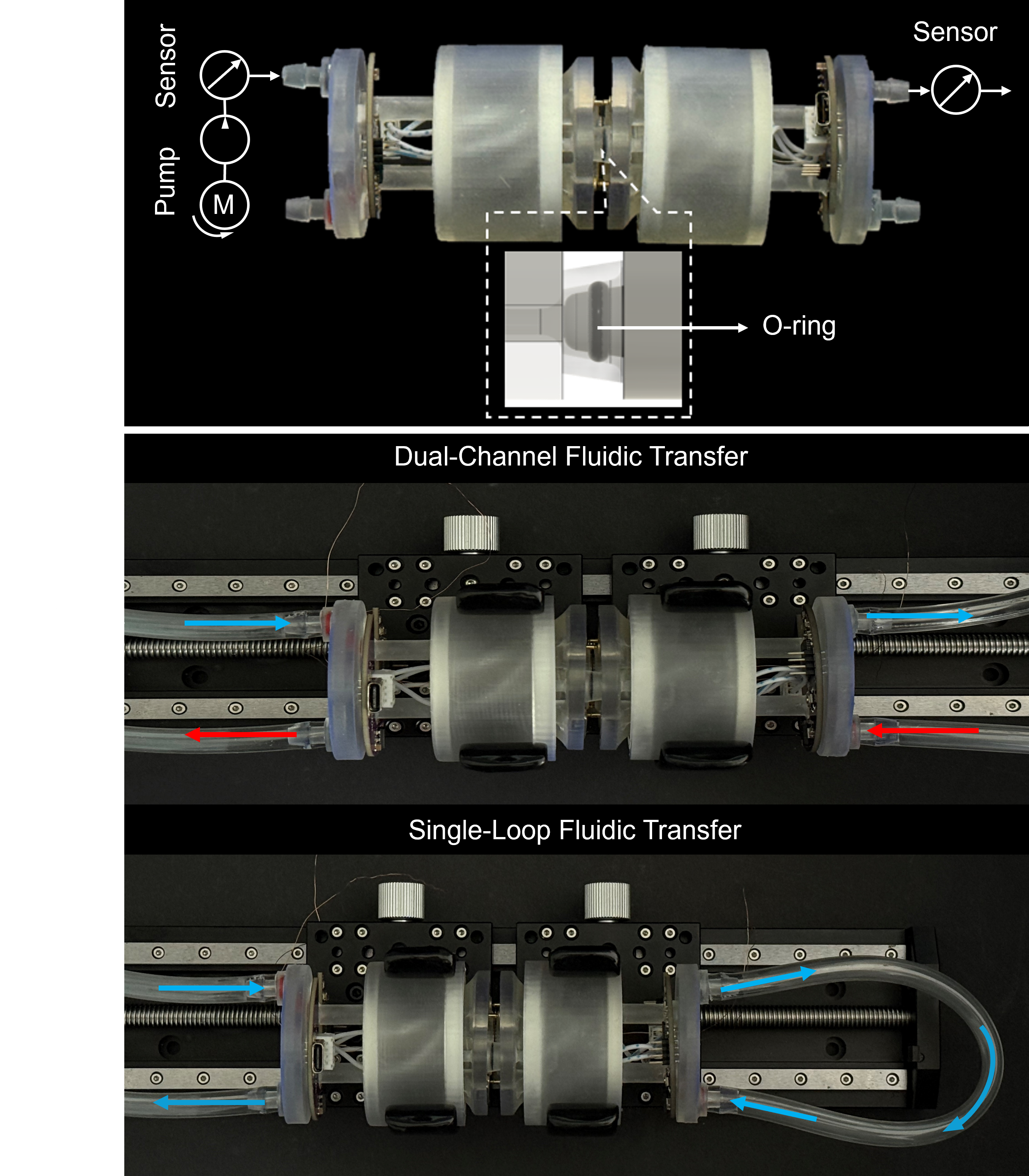} 
    \caption{Diagram of EPM connector configuration and fluid transfer experiment.}
    \label{fig:your_label}
\end{figure}
Analysis of the single-loop fluidic transfer data revealed consistent trends across different input flow rates. At an input of 80 ml/min, the outlet flow rate averaged approximately 49 ml/min, yielding an efficiency of about 61\% (Fig.~13a). Increasing the input to 90 ml/min resulted in a proportional rise in outlet flow rates to around 56 ml/min, maintaining efficiency at roughly 62\% (Fig.~13b). At the highest tested input of 100 ml/min, the outlet flow rate stabilized near 65 ml/min, corresponding to an efficiency close to 65\% (Fig.~13c). Variations in efficiency were minor across the tested input flow rates, suggesting stable connector performance under these conditions. The relatively lower efficiency observed in the single-loop mode can be attributed to the fluid passing through the connector twice. Consequently, the fluid traverses the water tank and internal channels twice as often, increasing energy losses and reducing overall efficiency.\\
\begin{figure}[htbp]
    \centering
    \begin{subfigure}[b]{0.5\textwidth}
        \includegraphics[width=\textwidth]{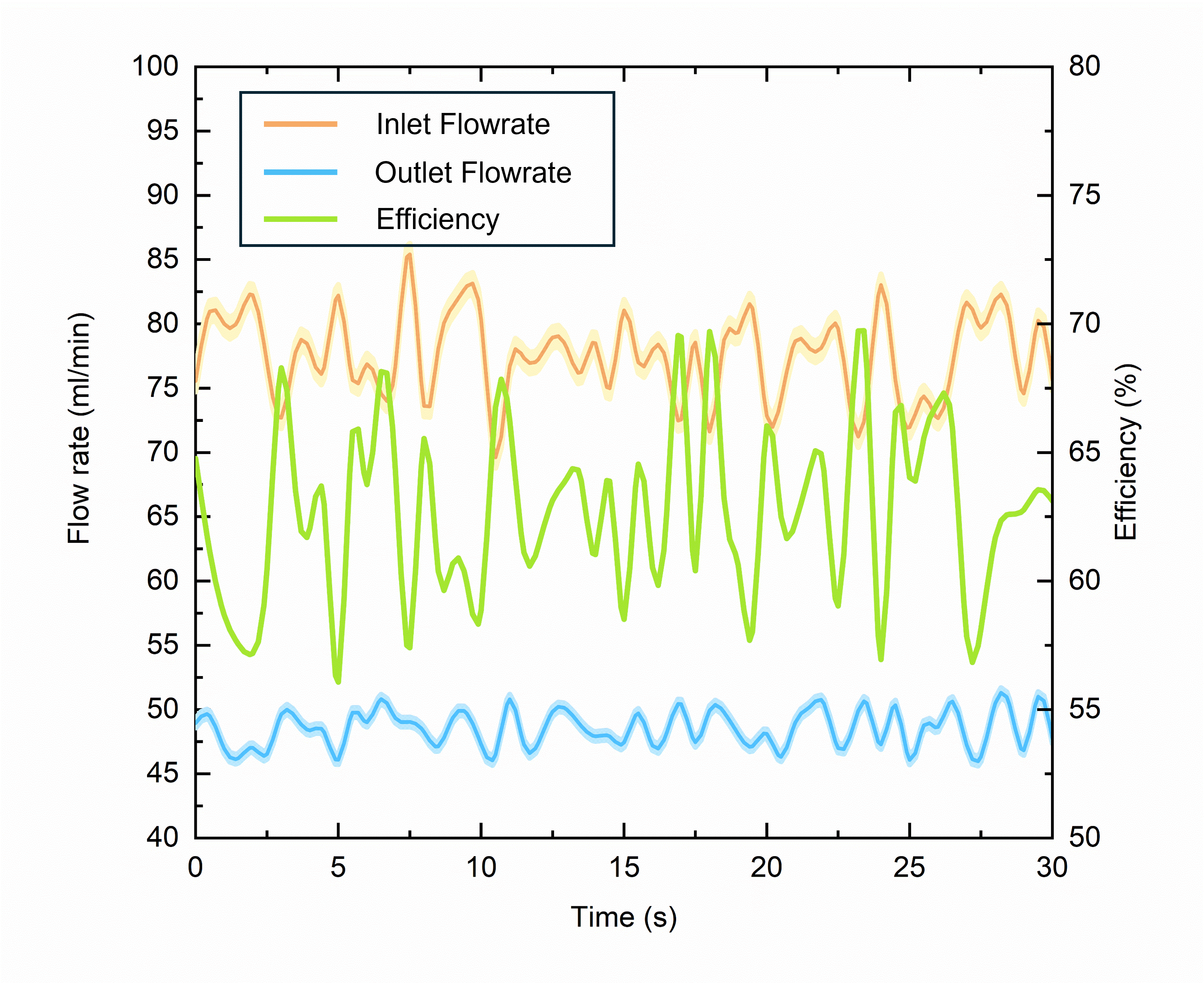}
        \caption{Flow rate and efficiency with 80 ml/min input
}
        \label{fig:sub1}
    \end{subfigure}
    \hfill
    \begin{subfigure}[b]{0.5\textwidth}
        \includegraphics[width=\textwidth]{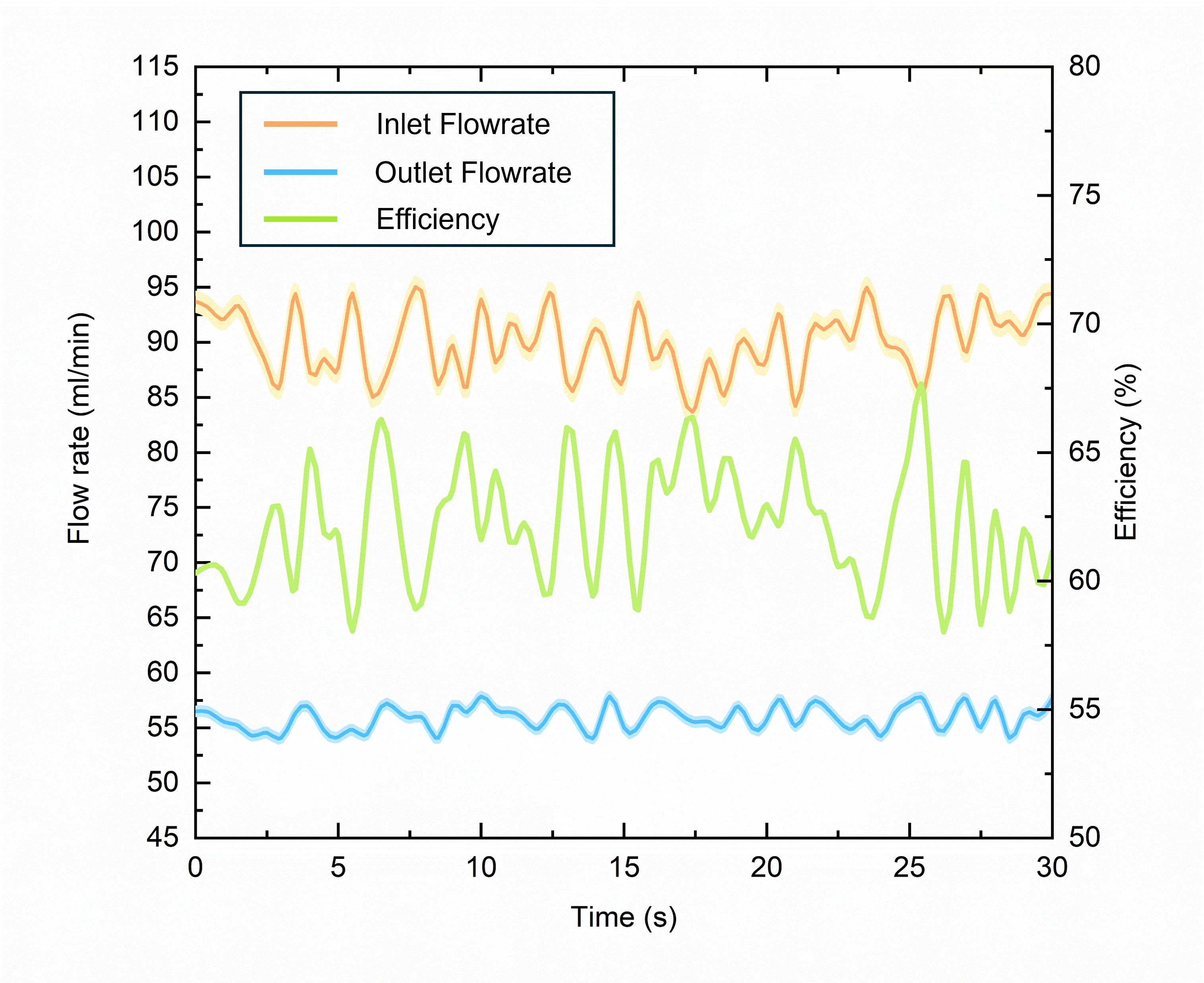}
        \caption{Flow rate and efficiency with 90 ml/min input
}
        \label{fig:sub2}
    \end{subfigure}
    \hfill
    \begin{subfigure}[b]{0.5\textwidth}
        \includegraphics[width=\textwidth]{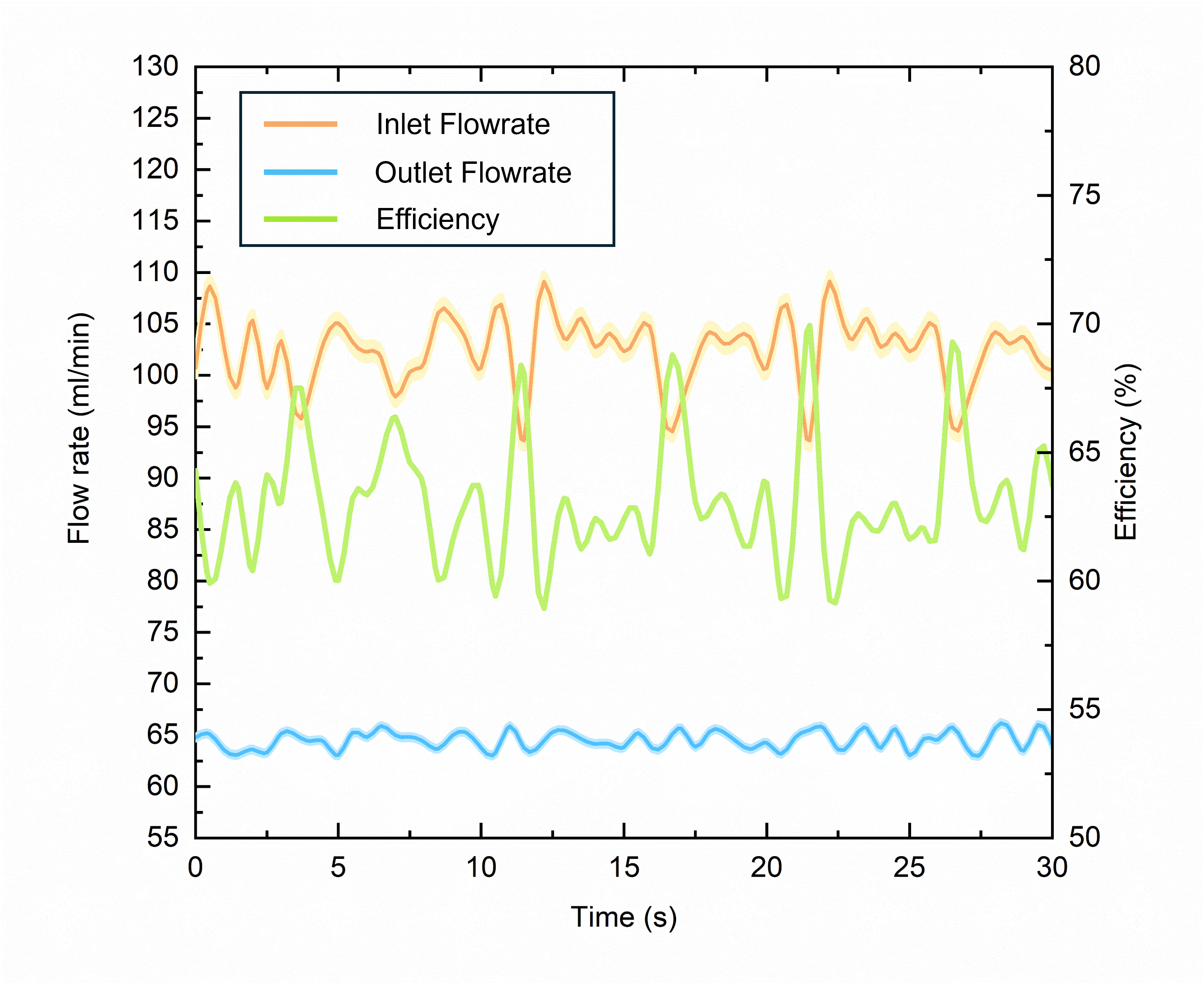}
        \caption{Flow rate and efficiency with 100 ml/min input
}
        \label{fig:sub3}
    \end{subfigure}
    \caption{Flow rate and efficiency data in single-loop fluidic transfer mode.}
    \label{fig:combined}
\end{figure}\\
The dual-channel fluidic transfer mode demonstrated significantly improved fluid transfer efficiency compared to the single-loop mode. At an input flow rate of 102 ml/min, the outlet flow rate closely matched at about 98 ml/min, resulting in an efficiency of around 95\% (Fig.~14a). Increasing the input to 140 ml/min led to a proportional rise in outlet flow rates to approximately 136 ml/min, maintaining efficiency at about 97\% (Fig.~14b). At the highest tested input of 175 ml/min, the outlet flow rate averaged near 170 ml/min, corresponding to an efficiency consistently around 98\% (Fig.~14c). These results indicate highly stable and efficient performance in the dual-channel fluidic transfer mode.\\
In summary, these experiments demonstrate that the dual-channel fluidic transfer mode significantly enhances fluid transfer efficiency compared to the Single-Loop mode. Both modes maintained stable performance across varying operational voltages, indicating reliable functionality of the connector under different operational conditions.

\begin{figure}[htbp]
    \centering
    \begin{subfigure}[b]{0.5\textwidth}
        \includegraphics[width=\textwidth]{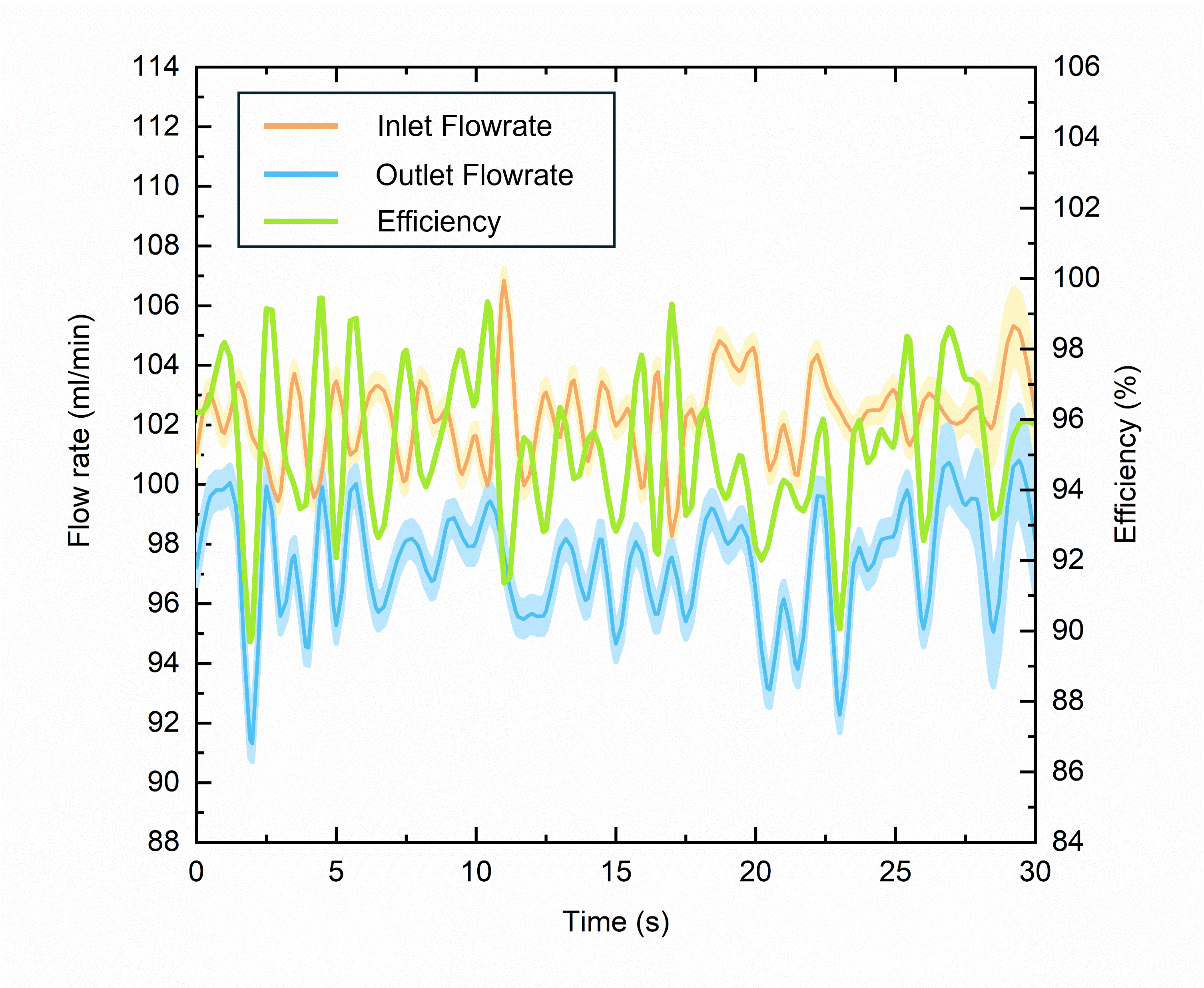}
        \caption{Flow rate and efficiency with 102 ml/min input
}
        \label{fig:sub1}
    \end{subfigure}
    \hfill
    \begin{subfigure}[b]{0.5\textwidth}
        \includegraphics[width=\textwidth]{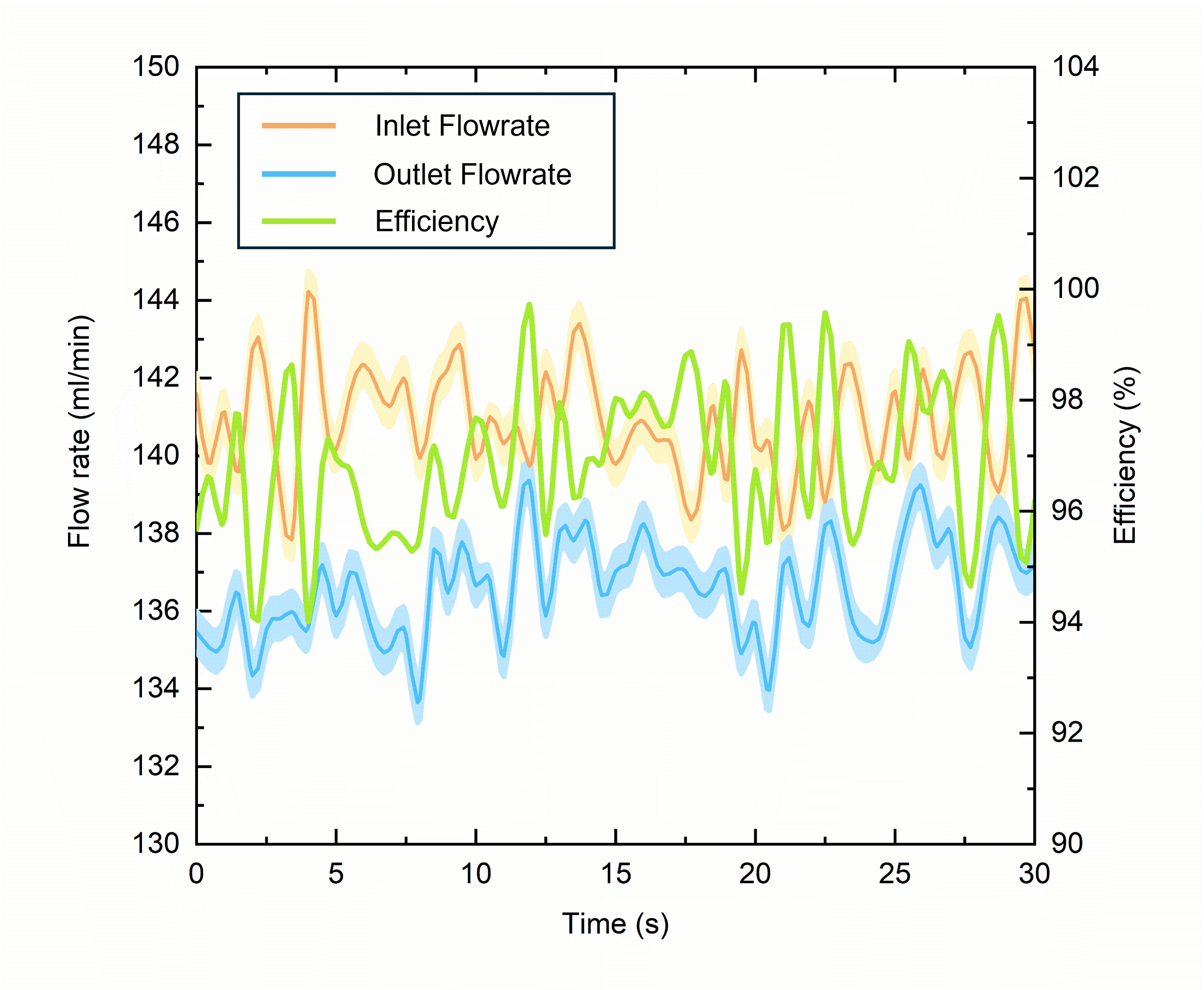}
        \caption{Flow rate and efficiency with 140 ml/min input
}
        \label{fig:sub2}
    \end{subfigure} 
    \hfill
    \begin{subfigure}[b]{0.5\textwidth}
        \includegraphics[width=\textwidth]{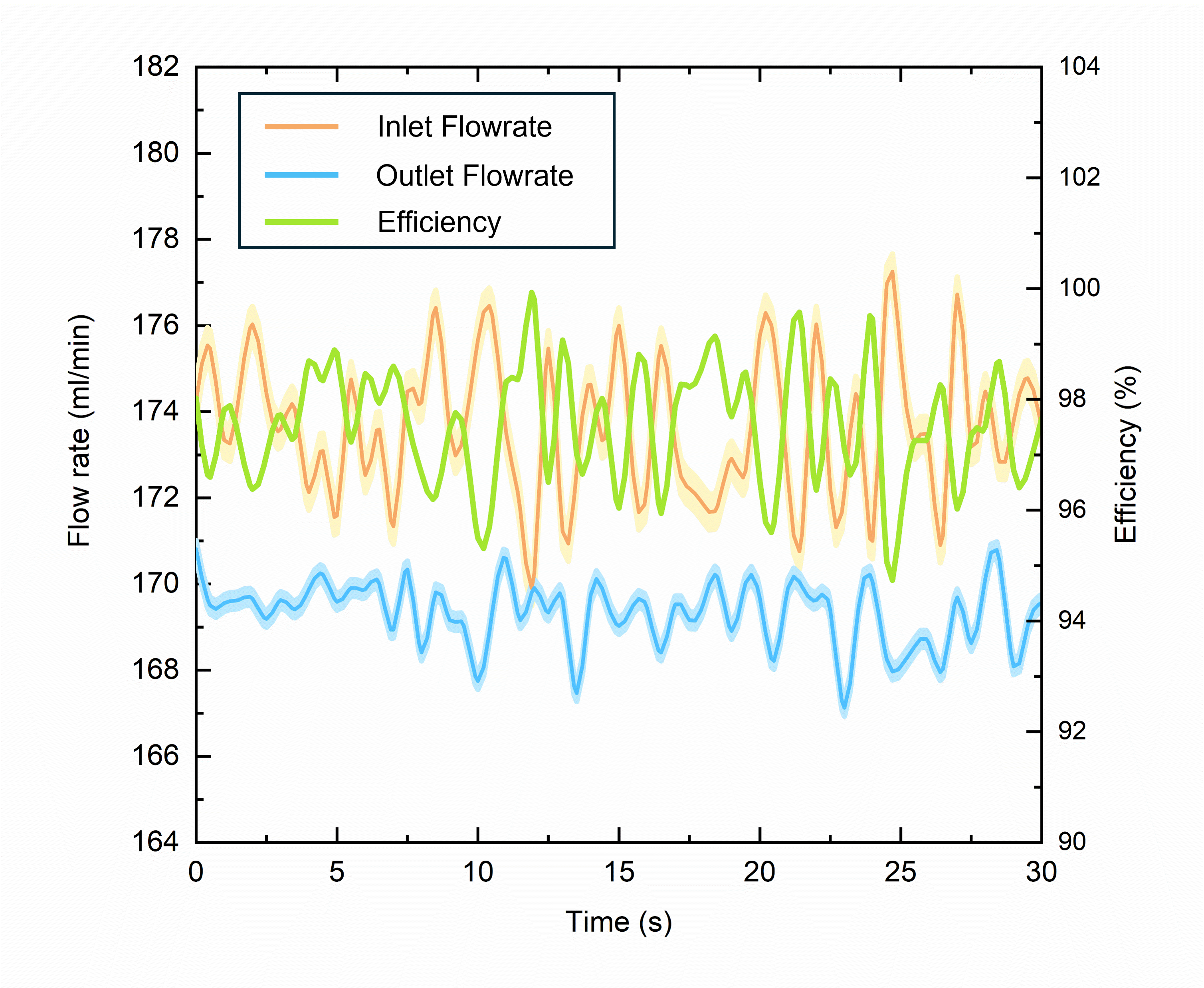}
        \caption{Flow rate and efficiency with 175 ml/min input
}
        \label{fig:sub3}
    \end{subfigure}
    \caption{Flow rate and efficiency data in dual-channel fluidic transfer mode.}
    \label{fig:combined}
\end{figure}

\subsection*{EPM connector for mechanical connection}
For applications primarily focused on mechanical connections, the electro-permanent magnet (EPM) connector provides an ideal solution by integrating flexible coupling, electro-permanent magnet technology, and magnetic self-alignment capabilities (Fig.~15). Structurally, the connector is divided into two primary sections—upper and lower—connected by a conventional compression spring.\\
The upper section consists of an upper housing, an internal rotor, an EPM, arc-shaped magnets, a bearing, an end cover, and embedded permanent magnets. A bearing facilitates the rotation of the internal rotor relative to the outer casing. The strategically positioned EPM and arc-shaped magnets within the internal rotor enable automatic magnetic alignment, significantly simplifying the coupling process. The lower section comprises a lower housing, a bottom rotor, another bearing, and additional embedded permanent magnets. These magnets interact with those embedded in the upper section, effectively connecting the upper and lower sections together and reinforcing secure alignment upon engagement.\\
A conventional compression spring mechanically links the upper and lower sections. In contrast to the conical spring employed in fluidic connectors, this conventional spring provides a simpler and potentially more flexible mechanical response, accommodating a greater range of positional variations and misalignments.\\
Upon activation, the EPM establishes a robust, reversible magnetic connection, allowing the spring to smoothly extend or retract depending on magnetic attraction or repulsion. Swift demagnetization returns the connector promptly to its original state, guaranteeing consistent and reliable mechanical alignment. This design allows rapid connection and disconnection without continuous current, greatly reducing energy consumption. Additionally, the bearings support relative rotational movement, and the spring’s compliance facilitates flexible alignment, making the connector particularly suitable for modular systems requiring adaptability and mechanical flexibility.

\begin{figure}[!t] 
    \centering
    \includegraphics[width=0.9\columnwidth]{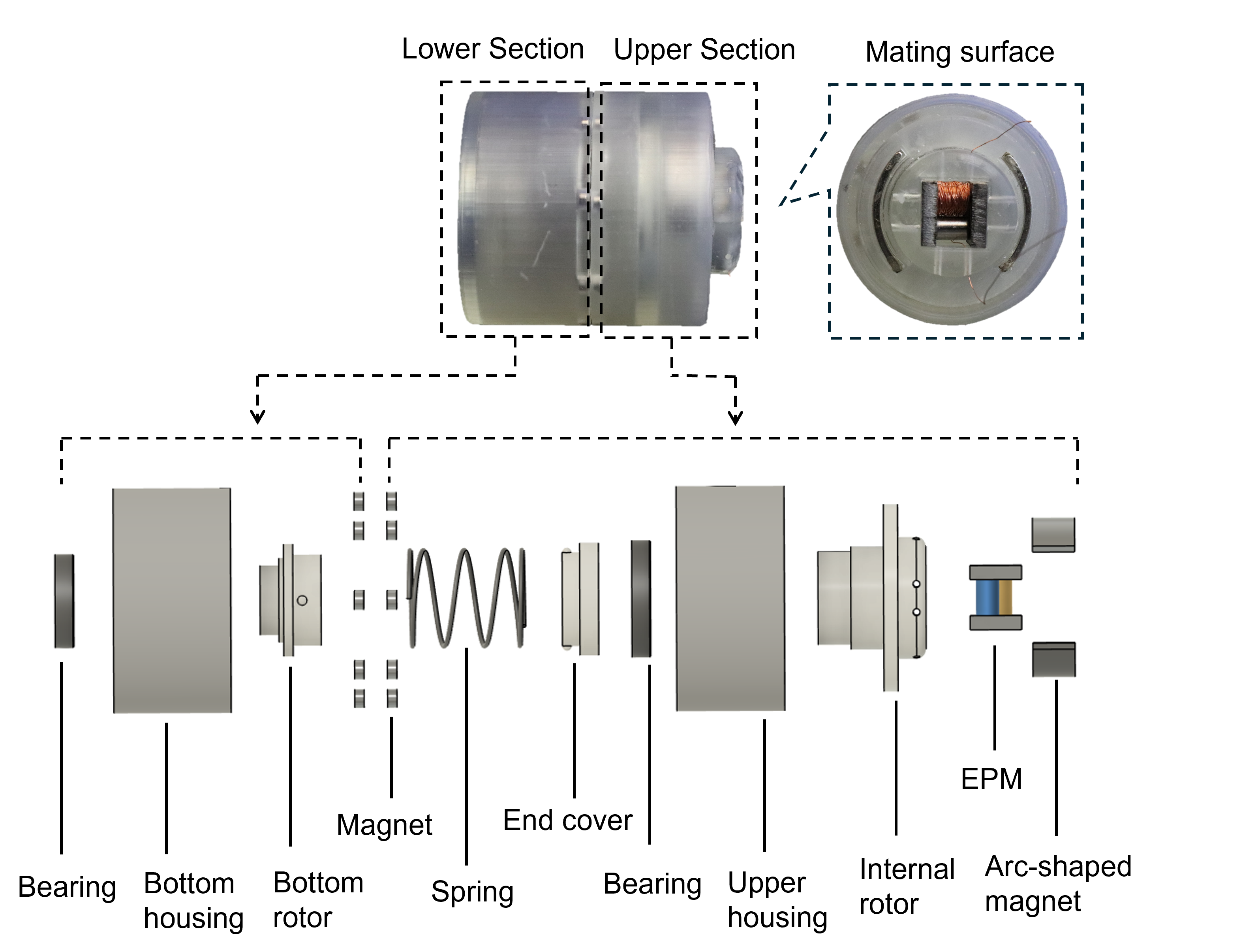} 
    \caption{Structure of EPM mechanical connector.} 
    \label{fig:your_label}
\end{figure}

\subsection*{Characterization of the EPM connector's mechanical coupling capability}
\subsubsection*{Mechanical Connection Flexibility}

To comprehensively characterize the mechanical coupling performance and flexibility of the connector, several tests were conducted (Fig.~16, Video S7). The axial extension test demonstrated the connector's capacity for substantial axial deformation, achieving a maximum extension of 20~mm when subjected to a gradually increasing tensile force. In the angular flexibility test, the connector accommodated a significant angular misalignment, reaching a maximum bending angle of approximately 30$^{\circ}$ when a bending moment was applied. Additionally, the lateral offset test confirmed the connector's tolerance to horizontal displacement, successfully maintaining coupling at an offset of up to 6~mm. Lastly, the maximum connection distance test evaluated the robustness of the connector under extended conditions, showing that connectors could remain securely coupled over distances of up to 132~mm. These results collectively highlight the connector’s substantial mechanical adaptability and robust coupling capability across various alignment scenarios.
\begin{figure}[!t] 
    \centering
    \includegraphics[width=0.8\columnwidth]{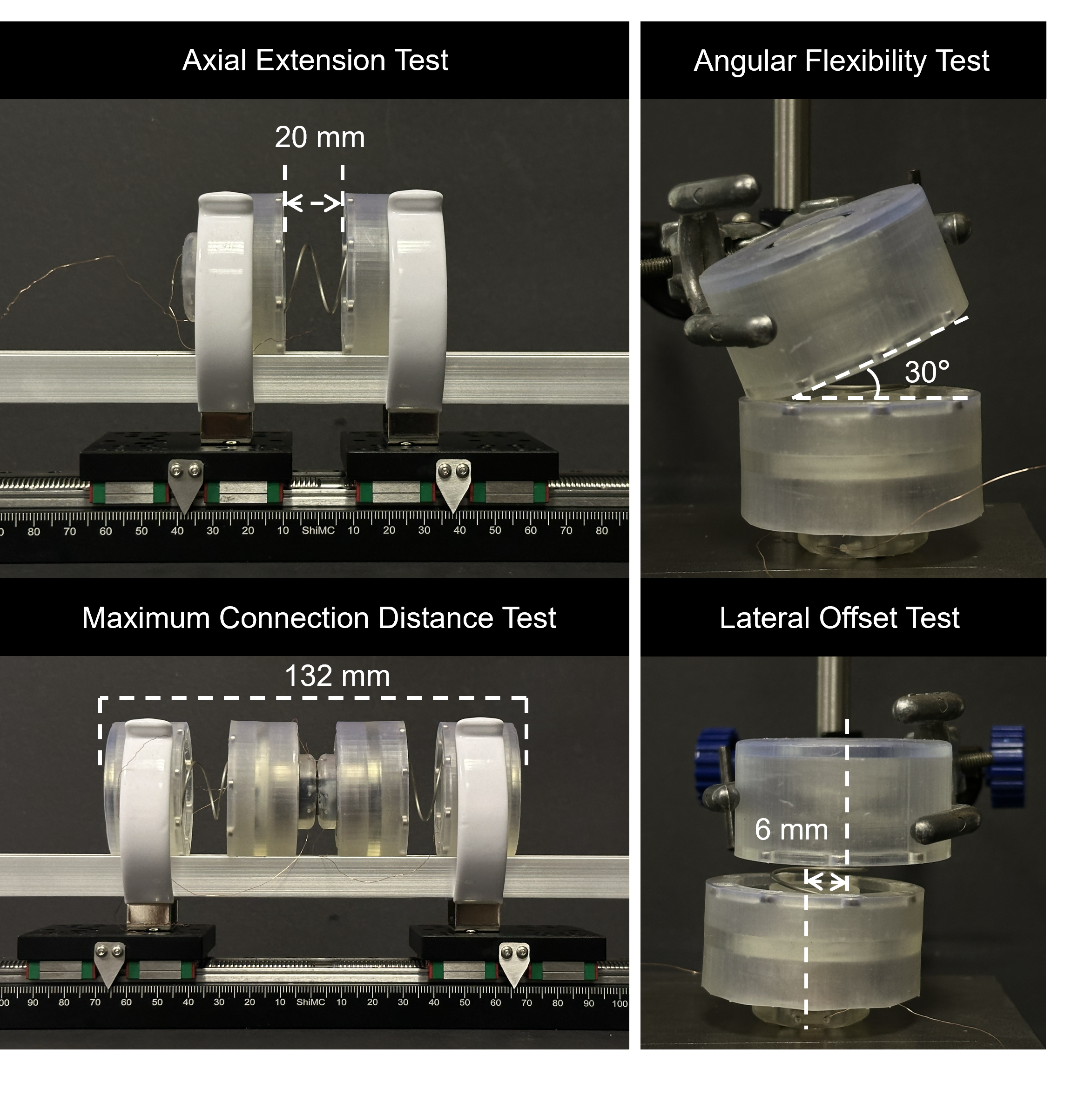} 
    \caption{EPM mechanical connector flexibility test.}
    \label{fig:your_label}
\end{figure}

\section*{DISCUSSION}

The multifunctional EPM connector significantly advances existing connector technologies by integrating mechanical coupling, self-alignment, efficient fluid transfer, and robust data communication into a single system. The use of electro-permanent magnet (EPM) technology enables rapid, reversible connections without continuous power consumption, greatly reducing energy demands, which is highly beneficial for modular robotics and autonomous applications. Additionally, the optimized coil configuration around the Alnico magnet enhances magnetic performance and energy efficiency, underscoring the critical role of precise alignment in maintaining coupling strength.\\
The connector’s mechanical adaptability, achieved through magnetic interactions, integrated bearings, and flexible springs, effectively manages significant axial, angular, and lateral misalignments. Its design minimizes energy consumption per switching event, resulting in notably low operational energy requirements. Moreover, the use of cost-effective SLA-based 3D printing technology significantly reduces manufacturing costs, promoting wider adoption and practical scalability. Fluid transfer tests confirmed the connector's capability, demonstrating high efficiency in dual-channel mode due to the internal isolation design, ensuring reliable, leak-free operation (Video S8).

\subsection*{LIMITATIONS AND FUTURE WORK}

Although the connector demonstrates notable performance improvements, several limitations remain to be addressed. Firstly, while fluid transfer performance is efficient for typical fluids such as water, challenges arise when handling high-density or particulate-containing fluids, potentially causing internal channel blockages and significantly reducing transfer efficiency. Secondly, the EPM activation relies on short, high-current pulses, necessitating robust external circuitry capable of delivering instantaneous high-energy demands.\\
Future research should explore improved channel designs or materials suitable for viscous and particulate-laden fluids. Additionally, integrating advanced sensors within the connector could enhance positional accuracy detection, thereby further improving self-alignment precision—an essential capability for autonomous and automated applications.

\section*{METHODS}


\subsection*{EPM connector design and fabrication}
The EPM connector was designed in Autodesk Fusion and 3D printed on an SLA 3D printer (Formlabs Form 3) using resin v4 clear material. Printed parts include the water tank,body shell,rotary joint, and mating surface (STL files of the components are available and described in more detail in the paper and supplemental information).\\
The tube inside the EPM connector is a silicone tube with a 3mm outer diameter and a 2 mm inner diameter.The dimensions of the two bearings are 35 × 44 × 5 mm and 25 × 32 × 4 mm, respectively.
\subsection*{EPM design and fabrication}
The EPM fabrication details are in the supplemental information.

\newpage


\section*{RESOURCE AVAILABILITY}


\subsection*{Lead contact}


Requests for further information and resources should be directed to and will be fulfilled by the lead contact, Adam A. Stokes (adam.stokes@ed.ac.uk).

\subsection*{Materials availability}


This study did not generate new unique reagents.

\subsection*{Data and code availability}


All data and source code underlying this study are available from the corresponding author upon reasonable request.

\section*{ACKNOWLEDGMENTS}


This research received no specific grant from any funding agency in the public, commercial, or not-for-profit sectors. The authors are grateful to the members of the lab for their constructive feedback and assistance throughout the study.

\section*{AUTHOR CONTRIBUTIONS}


Conceptualization, B.W.; methodology, B.W.; writing—original draft, B.W. and A.A.S.; writing—review and editing, A.A.S.; supervision, A.A.S.

\section*{DECLARATION OF INTERESTS}


The authors declare no competing interests.

\section*{DECLARATION OF GENERATIVE AI AND AI-ASSISTED TECHNOLOGIES}

During the preparation of this work, the authors used ChatGPT-4o (OpenAI) to enhance the clarity and language quality of the manuscript. The tool was also used to assist in drafting the initial description of the experimental objective associated with Figure 10. All AI-assisted content was carefully reviewed, revised, and validated by the authors, who take full responsibility for the integrity and accuracy of the final publication.

\section*{SUPPLEMENTAL INFORMATION INDEX}




\begin{description}
  \item Figure S1. EPM fabrication steps.
  \item Figure S2. Pulse vs. magnetic field for the EPM.
  \item Figure S3. Pulse response of the EPM coil current (green) to a 1 ms voltage pulse (blue).
  \item Figure S4. Magnetic field lines between two arc-shaped magnets (blue: north pole, red: south pole)
  \item Figure S5. EPM holding force test experimental setting.
  \item Figure S6. Histogram of holding forces under a normal load.
  \item Figure S7. EPM connector inside section.
  \item Figure S8. PCB schematic.

  \end{description}

\newpage

\bibliography{article-template.bib}

\bigskip


\newpage

\end{document}